# Enhancing Hindi NER in Low Context: A Comparative study of Transformer-based models with vs. without Retrieval Augmentation

**Sumit Singh**
Indian Institute of Information Technology, Allahabad
sumitrsch@gmail.com

**Rohit Mishra**
Indian Institute of Information Technology, Allahabad
rohit.129@gmail.com

**Uma Shanker Tiwary**
Indian Institute of Information Technology, Allahabad
ust@iiita.ac.in

July 21, 2025

## Abstract

One major challenge in natural language processing is named entity recognition (NER), which identifies and categorises named entities in textual input. In order to improve NER, this study investigates a Hindi NER technique that makes use of Hindi-specific pretrained encoders (MuRIL and XLM-R) and Generative Models ( Llama-2-7B-chat-hf (Llama2-7B), Llama-2-70B-chat-hf (Llama2-70B), Llama-3-70B-Instruct (Llama3-70B) and GPT3.5-turbo), and augments the data with retrieved data from external relevant contexts, notably from Wikipedia. We have fine-tuned MuRIL, XLM-R and Llama2-7B with and without RA. However, Llama2-70B, lama3-70B and GPT3.5-turbo are utilised for few-shot NER generation. Our investigation shows that the mentioned language models (LMs) with Retrieval Augmentation (RA) outperform baseline methods that don't incorporate RA in most cases. The macro F1 scores for MuRIL and XLM-R are 0.69 and 0.495, respectively, without RA and increase to 0.70 and 0.71, respectively, in the presence of RA. Fine-tuned Llama2-7B outperforms Llama2-7B by a significant margin. On the other hand the generative models which are not fine-tuned also perform better with augmented data. GPT3.5-turbo adopted RA well; however, Llama2-70B and llama3-70B did not adopt RA with our retrieval context. The findings show that RA significantly improves performance, especially for low-context data. This study adds significant knowledge about how best to use data augmentation methods and pretrained models to enhance NER performance, particularly in languages with limited resources.

## 1 Introduction

Named entity recognition (NER) [1], is a natural language processing task that involves identifying and categorizing named entities in structured and unstructured text, such as people, organizations, locations, dates, and other important information.

For instance, a Hindi sentence is given: **"इसे ऑस्ट्रेलियाई रिकॉर्डिंग उद्योग संघ द्वारा ७०,००० से अधिक प्रतियों की बिक्री के लिए प्लेटिनम प्रमाणित किया गया था। "** Corresponding English sentence: "It was certified platinum by the Australian Recording Industry Association for sales of over 70,000 copies."

For this sentence, the NER model identifies "ऑस्ट्रेलियाई रिकॉर्डिंग उद्योग संघ" ( "Australian Recording Industry Association" ) as Corporation.

Named entity recognition is used in various applications, such as knowledge graph generation [2], question answering [3, 4], text summarization [5], machine translation [6, 7], Web Scraping and information retrieval [8].

The low-context example has been defined as one with a smaller context for describing an entity. Simplest examples [10] are examples which have smaller lengths, like one, two, three or less than 5. Our focus is on low context entities, which are also among many challenges listed in [10]. For low context entities, methods for integrating external knowledge, e.g., Knowledge Bases (KBs) [11] or Gazetteer [10], into neural architectures, has gained renewed attention. A Gazetteers-based neural network model used by [12] to overcome the low context challenges, but it requires domain-specific Gazetteers.

[11] gained renowned attention for using Wikipedia knowledge-base as Retrieval Augmented approaches by getting the state-of-the-art result for Multiconer-1 data [13]; however, [11] utilised a multi-stage training approach, incorporating XLM-R Language Model (LM) with a CRF classification layer, and utilized the Retrieved Augmentation approach across multiple language datasets. It ensemble the top models for best results and achieved 86.23 macro F1 score for the Hindi language.

Our work is motivated by the [11] and focused on finding the contribution of Retrieval augmentation (RA) with Hindi-specific LMs (XLM-R [16], Muril [15]) and generative LMs (Llama2-7B, Llama2-70B Llama3-70B (series of [36]) and GPT3.5-turbo (optimized version of [35])) at the low context data for the Hindi language. An example of our work is shown in Fig. 1, which depicts the effect of RA by the MuRIL model.

To investigate the effect of the RA in Hindi, transformer-based encoders [14], MuRIL [15] and XLM-R [16] pretrained models selected for fine-tuning the task since in [17, 18] it is reported that these two models have performed better than other models for NER task. To the best of our knowledge, RA has not been applied using MuRIL for the Hindi NER task. We have also fin-tuned the llama2-7 for the NER task in both the settings with and without RA. Also, generative models (Llama-2-7, Llama-2-70, Llama-3-70and GPT3.5-turbo) were selected for the experiment in which Llama series is open source, and GPT3.5-turbo requires a subscription. Results of MuRIL [15] and XLM-R [16] without KB is tabulated in first row of Table 7 [17]. GPT3.5 turbo was selected based on its performance in [29]

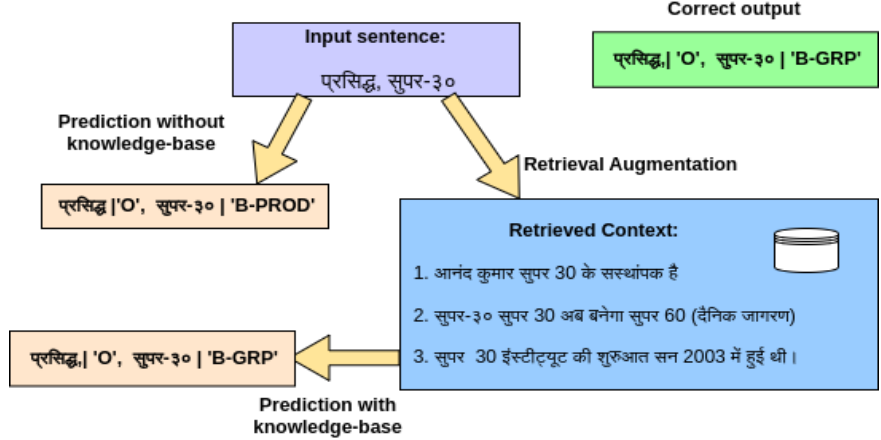

Figure 1: Demonstration of our Model for an example of test data.

In implementing RA, we constructed a knowledge base (KB) using the Wiki dump and utilized similar contexts to augment the examples, as described in [11]. The process of RA is described in Section 4.1.1.

Our contributions:

1. Fine-tuned Llama2-7B model for NER task using Hindi data with and without retrieval augmentation. Results indicate that Llama2-7B produces better accuracy and output format after finetuning than the base Llama2-7B model.
2. Performed few-shot-based prompt engineering with LLMs such as Llama-2-70B-chat-hf (on twenty thousand test examples), Llama-3-70B-Instruct (on twenty thousand test examples) and GPT3.5-turbo (on one thousand test examples) for Hindi NER task with and without Retrieval Augmentation (RA).

3. Performed model fine-tuning with (XLM-R and MuRIL) at the Hindi NER dataset with and without RA.
4. We have compared the effect of RA on the performance transformer-based encoders (MuRIL and XLM-R ) and Large Language Models (Llama-2-70B-chat-hf, Llama-3-70B-Instruct and GPT3.5 turbo LLMs ) for the Hindi NER task. Our analysis indicates that when examples are enhanced through RA, NER models based on transformer-based LMs exhibit enhanced performance, specifically for low context examples, as the macro F1 score for the XLM-R model rose from 0.49 to 0.71, while the macro F1 score for the MuRIL model increased from 0.69 to 0.70. Similarly, macro F1 score for the GPT3.5-turbo model increased from 0.17 to 0.31. However, Llama2-70B and Llama3-70B did not show any improvement with RA, and the reason may be a shorter context window.

Including this section, this paper is divided into seven sections. Section 2 discusses the previous work done related to this task. Section 3 explains the dataset that is used in this paper. Section 4 describes the methodologies for three approaches, which include the process of RA, selection of LMs, NER module and details of the experiment setup. Section 5 gives brief information about evaluation metrics. Section 6 discusses the results obtained by using the RA with both models and the analysis of errors. Finally, in section 7, the paper concludes.

## 2 Related work

Deep Learning has significantly enhanced the performance of Named Entity Recognition (NER) in recent times. Using neural approaches, such as BiLSTM-CRF models [19, 20], which employ static word embeddings, has eliminated the need for manual feature engineering, such as capitalization features. More recently, Transformer-based LMs like Bert [21] have shown competitive results in various NLP tasks, including NER. However, these models face challenges in handling low-context situations, long-tail entities, emerging entities, and complex entities as described in [10] since they generate contextual word embeddings and are pre-trained only on existing data.

To address the issue of emerging entities in web queries within a multi-lingual code-mixed context, the approach presented in [22] offers a solution. Similar challenges are also encountered in [23] and HiNER [24]. Both [22] and [10] propose integrating a Gazetteer with language models to solve these challenges. However, the use of a Gazetteer poses certain challenges. For instance, the Gazetteer should cover entities in a sufficient ratio, and for new domains and entity types, a specific Gazetteer is required, which depends on its availability. Furthermore, as stated in [10], a weak learning problem arises when the model already knows the entity type in advance for entities covered by the Gazetteer.

One prominent approach to address the problem of lower context discussed in [23] is based on a Knowledge-based System [11] that augments text from Wikipedia with the NER data to handle examples with lower context, it also utilizes XLM-R pretrained model for learning. Additionally, [12] achieves good results by integrating language models like BERT with a Gazetteer network, minimizing the KL divergence. Both [25] and [11] employ multi-stage training to leverage the advantages of multi-lingual tracks. Moreover, [25] utilizes an Ensemble architecture and data augmentation to create a more robust model suitable for noisy environments. [17] fine-tunes MuRIL-large without any knowledge base at Multiconer-1 data and achieves F1 scores of 0.69 and 0.59 for Hindi and Bangla languages, respectively. [24] provides another NER dataset for the Hindi language, which benchmarks XLM-R, MuRIL, m-bert [21] and Indic-bert [26] with a cross-entropy classification layer, and reported that XLM-R and MuRIL performed better. The zero-shot generalisation of large language models (LLMs) [27] has also revolutionised natural language processing (NLP). An analysis of zero-shot NER using ChatGPT was conducted in [28, 29, 30], and so far, the findings were not as good as the Benchmark results using the transformer encoder-based models. Fine-tuning generative models is expensive; however, QLoRA [39] has reduced the cost to some extent with minor quality degradation.

## 3 Data

Multiconer-1 data [13] have arranged 13 tasks for different languages. We selected the Hindi language for our experiment. There are 15300 samples in training and 800 in development for Hindi tracks. The Hindi testing dataset has 141565 samples with a total of 933273 tokens. The motivation behind selecting this dataset is its testing data, which has a large number of examples (141565), and a huge amount of low-context examples. It contains six types of named entities for the Hindi language, which include Location (LOC), Person (PER), Production (PROD), Group (GRP), Corporation (CORP), and Creative Work (CW). The dataset is in CONLL format, which uses BIO tagging to label the named entities. Each dataset example has a sequence of words (sentence) and a corresponding sequence of labels. Table 1 presents the number of each entity in training, development (validation) and testing datasets.

| Tag | Training | Validation | Testing |
|---|---|---|---|
| B-LOC | 2614 | 131 | 31546 |
| B-PER | 2418 | 133 | 25353 |
| B-PROD | 3077 | 169 | 22399 |
| B-GRP | 2843 | 148 | 22140 |
| B-CORP | 2700 | 134 | 21713 |
| B-CW | 2304 | 113 | 21789 |
| I-LOC | 1604 | 77 | 25442 |
| I-PER | 2836 | 166 | 31197 |
| I-PROD | 2295 | 107 | 15839 |
| I-GRP | 5821 | 297 | 43930 |
| I-CORP | 2917 | 138 | 26762 |
| I-CW | 3592 | 151 | 37639 |
| O | 209545 | 10882 | 607524 |
| Total | 244566 | 12646 | 933273 |

Table 1: Entity distribution for the dataset.

In the testing dataset, the number of sentences whose length is less than 5 is pretty large. Fig. 2 displayed the graph of the length of sentences versus the number of occurrences in each data set, and the numerical value tabulated in Table 2. It is evident from this table that the sentences with lengths of two, three, four, or five have examples of more than 12000 each; therefore, these examples have lower context.

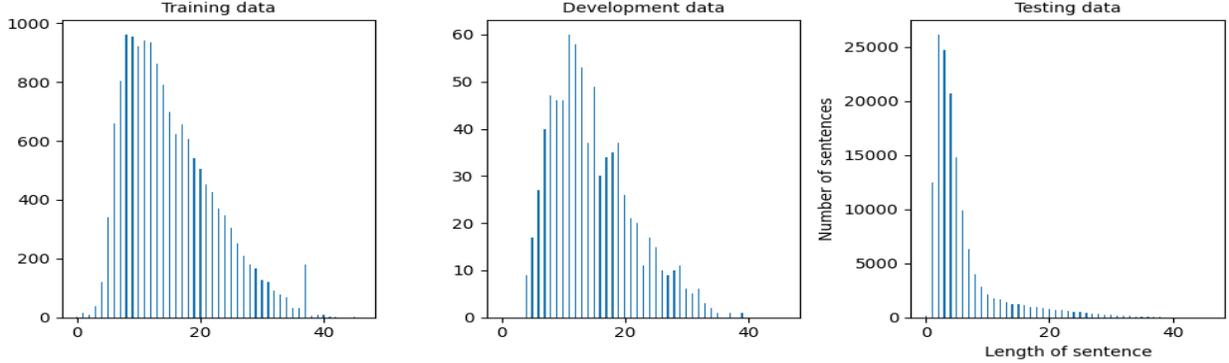

Figure 2: Length-wise distribution of the dataset.

| Length of examples | 1 | 2 | 3 | 4 | 5 | 6 | 7 | 8 |
|---|---|---|---|---|---|---|---|---|
| Number of examples | 142 | 12470 | 26159 | 24721 | 20727 | 14807 | 9926 | 6326 |
| Length of examples | 9 | 10 | 11 | 12 | 13 | 14 | 15 | 16 |
| Number of examples | 4011 | 2863 | 2089 | 1752 | 1678 | 1427 | 1237 | 1209 |
| Length of examples | 17 | 18 | 19 | 20 | 21 | 22 | 23 | 24 |
| Number of examples | 1125 | 990 | 959 | 894 | 806 | 705 | 656 | 636 |
| Length of examples | 25 | 26 | 27 | 28 | 29 | 30 | 31 | 32 |
| Number of examples | 503 | 468 | 464 | 358 | 319 | 253 | 201 | 163 |
| Length of examples | 33 | 34 | 35 | 36 | 37 | 38 | 39 | 40 |
| Number of examples | 124 | 112 | 83 | 66 | 51 | 33 | 25 | 6 |
| Length of examples | 41 | 42 | 43 | 44 | 45 | 46 | 47 | 48 |
| Number of examples | 10 | 5 | 2 | 2 | 1 | 0 | 1 | 0 |

Table 2: Length-wise distribution of Examples in test dataset

## 4 METHODOLOGY

This section is divided into three parts, The first part describes Retrieval Augmentation fine-tuning (RAFT) with XLM-R and MuRIL. The second part describes RAFT with the Llama2-7B. The third part shows the performance with Llama2, Llama3, and GPT3.5 turbo without fine-tuning for the NER task.

### 4.1 RAFT with XLM-R and MuRIL

This section is further divided into four parts. First, we describe how a query (input data) is processed to retrieve context from the Wikipedia knowledge base and how the augmentation is done. The second part deals with selecting LMs, the third part describes the architecture for NER, and the fourth part discusses about experimental setup.

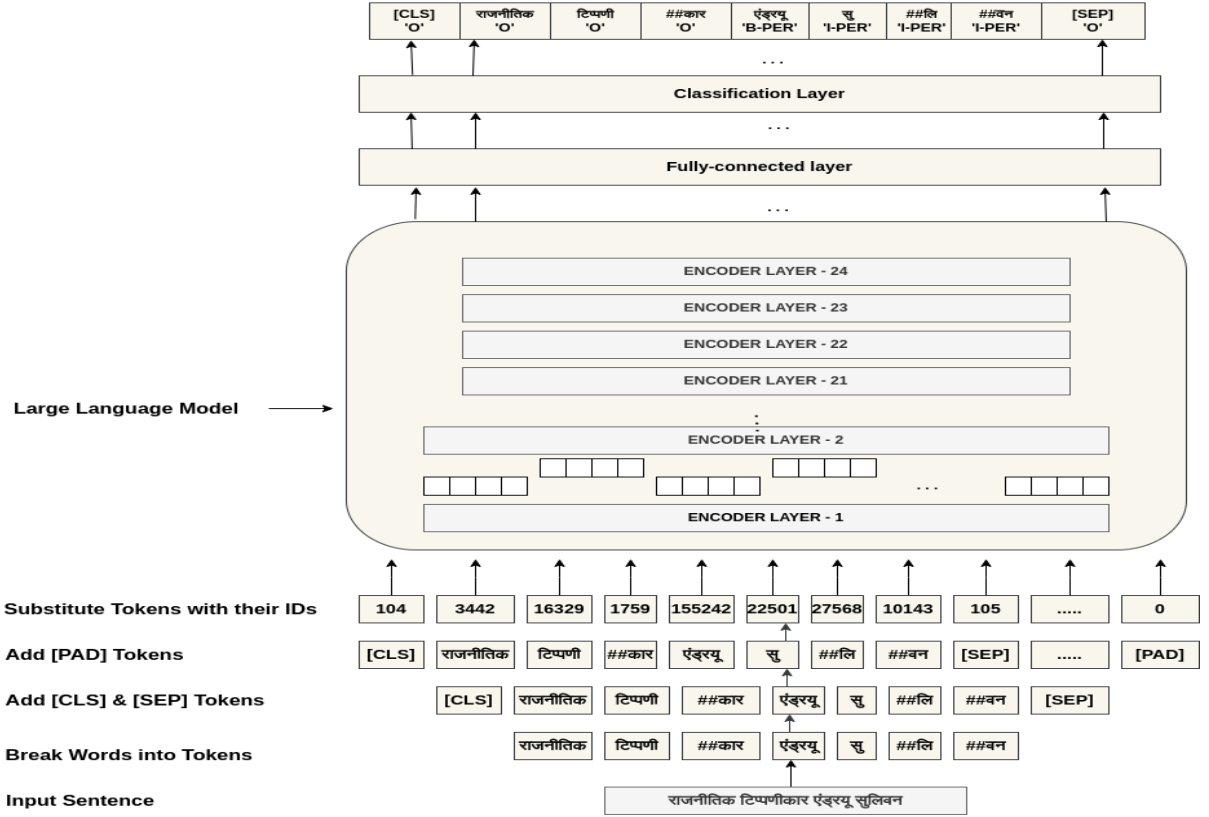

Figure 3: Generalized Transformer-based NER Model Architecture

#### 4.1.1 Retrieval Augmentation

Retrieve augmented context allows the inclusion of external contextual information to assist in disambiguating intricate named entities. Getting in more information through augmentation leads to a greater range of data that is used for getting an accurate understanding of context. For given input " विकी बर्मी साहित्य " augmented text are " विकी बर्मी साहित्य <EOS> [ बर्मी साहित्य ] इस युग को बर्मी साहित्य का स्वर्णकाल कहा जाता है। [ बर्मी साहित्य ] यह बर्मी साहित्य के विस्तार और प्रसार का युग है। [ बर्मी साहित्य ] बर्मी राजाओं द्वारा त्रिपिटक का अधिक अध्ययन होने से बर्मी साहित्य पर पालि का अत्यधिक प्रभाव पड़ने लगा। [ बर्मी साहित्य ] <e: हिंदी साहित्य>हिंदी साहित्य </e> में संत कवियों की तरह भिक्षुओं ने बर्मी साहित्य पर आधिपत्य कर लिया है। [ बर्मी साहित्य ] इस युग में बर्मी साहित्य की उन्नति पगन् युग से अधिक हुई। [ बर्मी साहित्य ] बर्मी साहित्य का अभ्युदय प्राय: काव्यकला को प्रोत्साहन देनेवाले राजाओं के दरबार में हुआ है इसलिए बर्मी साहित्य के मानवी कवियों का संबंध वैभवशाली महीपालों के साथ स्थापित है। [ बर्मी साहित्य ] बर्मी साहित्य के अंतर्गत बुद्धवचन (<e:त्रिपिटक>त्रिपिटक</e>), अट्टकथा तथा टीका ग्रंथों के अनुवाद सम्मिलित हैं। [ बर्मी साहित्य ] इस आज तक बर्मी साहित्य में एक प्रसिद्ध रचना के रूप किया जाता है। [ बर्मी साहित्य ] <e: बर्मी भाषा>बर्मी भाषा [ बर्मी साहित्य ] बर्मी भाषा में गद्य और पद्य दोनों प्रकार की साहित्य विधाएँ मौलिक रूप से मिलती हैं। "

This section is divided into two parts. The first part defines how the context for input examples is retrieved (Knowledge Retrieval), and the second part defines the process of augmentation (Data Augmentation).

1. **Knowledge Retrieval** This module is similar to [11]. The following paragraph explains this module in short.

   **Knowledge Base Building :**

   This work utilizes Hindi Wikipedia as the foundation to construct localized Wikipedia search engines, enabling us to retrieve the context of the input sentences. We download the Wikipedia dump[1] and convert it to plain text. After that, ElasticSearch[2] is used to index them. Wikipedia search engines define the document using three fields: sentence, paragraph, and title. This work constructs inverted indexes for both the sentence and title fields to facilitate efficient retrieval. The sentence field serves as a comprehensive text retrieval field, operating at the sentence level. In contrast, the title field identifies the entity addressed in the wiki page, making it suitable for entity-level retrieval. The contexts of the sentences are stored in the paragraph field.

   **Sentence Retrieval :** To obtain pertinent information at the sentence level, we employ the input example as a query and retrieve the most relevant documents based on the sentence field, selecting the top-k results. k is 10 in this work.

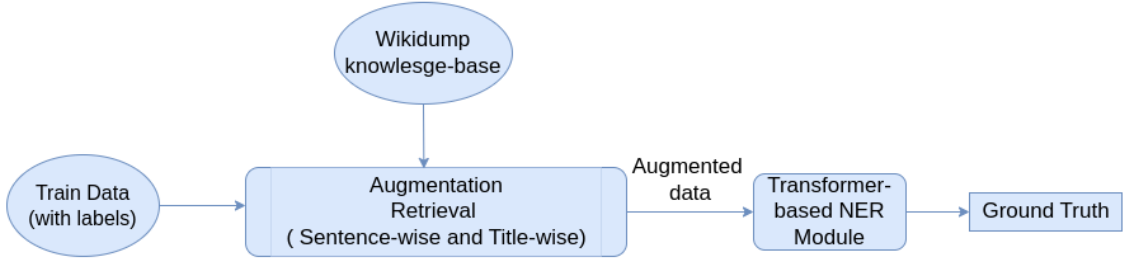
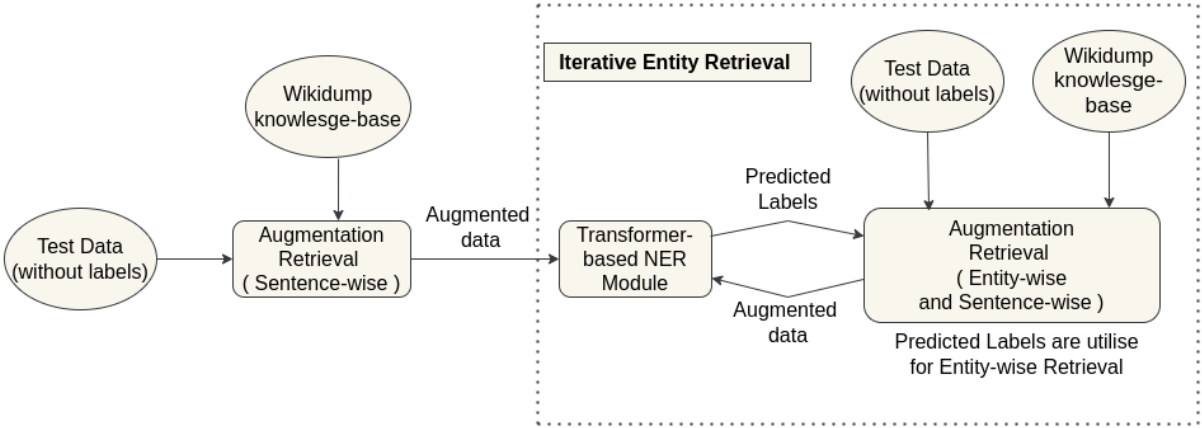

Figure 4: Figure shows an overview of the process of the training and testing phase. The training phase has labelled examples; therefore, training has been done with the augmented training data ( with sentence retrieval and title retrieval). The testing phase initially has examples without the label; therefore, examples are augmented with sentence retrieval for prediction at the first iteration. After the first iteration, sentence-wise and title-wise, iterative entity retrieval is done with the predicted entities of the previous iteration.

---

[1]https://dumps.wikimedia.org/hiwiki/latest/hiwiki-latest-pages-articles.xml.bz2
[2]https://www.elastic.co/

**Iterative Entity Retrieval :** The NER task prioritizes entities, but retrieving at the sentence level disregards the crucial entities present in the sentences. When retrieving documents, we search for the entities present in the query (sentence) in the title field of the database to address this issue.

This process utilises the ground-truth entities for the training and development datasets for entity retrieval. However, for the test dataset a two-step process is followed since in test data labels are not defined. Firstly, we perform sentence retrieval or leverage prior knowledge of the entity mentioned in the input sentences. Then, we augment the sentences and input them into the NER model for prediction. Now, based on the predicted labels, we find the entities which are present in the sentences irrespective of whether they are correct or not. Now again, we augment input sentences with sentence retrieval and updated entity retrieval so that with the help of updated knowledge, the NER model can know more about the input sentences. Hence, it is Iterative Entity retrieval. Iterations are performed based on improvement. The process of Iterative Entity retrieval is illustrated in Fig. 1

**Context Processing :** After retrieving the context documents (wiki page) from the input query, the context is generated using matching paragraphs from the top 10 retrieved documents. All ten generated contexts are concatenated to form the final context.

2. **Data Augmentation** According to Fig. 2, many sentences in the training, valid and testing data are less than five words. For training, we augmented each example of training and valid data to utilize the information of the augmented context. The retrieved data from the 4.1.1 was utilized for augmentation.

   An example of data augmentation for a low context test data from the original dataset (" विकी बर्मी साहित्य ") is shown in section 4.1.1. Here " विकी बर्मी साहित्य " is a test sentence, and "<EOS>" is a symbol which separates original data and augmented data. "[ बर्मी साहित्य ]" is the title of the page that is searched sentence-wise through input. "<e:हिंदी साहित्य>हिंदी साहित्य</e>" shows the hyperlinks as entities which are derived from the retrieved paragraph. A special symbol surrounds hyperlinks so the model can easily understand that they are hyperlinks to a title related to the original sentence. All retrieved paragraphs are separated with a separate special token of the model, e.g. "</s>" for XLM-R and "[SEP]" for MuRIL. Every word in the augmented portion is designated with a new label, "B-X". Augmented examples were fed into the language model for training. For the augmentation of the test data, we performed iterative entity retrieval as defined in 4.1.1. In the first iteration, only sentence retrieval is done as, at this stage, the entities present in the test examples are not known. After the first iteration of retrieval, our trained language model predicts the labels. The predicted labels are again utilised for augmentation retrieval. The predicted labels from the previous iteration were used in the next iteration for entity retrieval. In order to find similar titles, the words predicted as labels are searched in the title field. Based on these maximum similar titles again, augmentation is repeated.

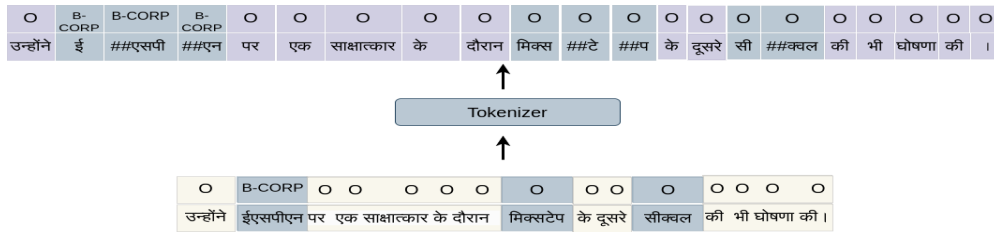

Figure 5: Some words are split into sub-word tokens during tokenization with MuRIL tokenizer. Labels are aligned accordingly.

### 4.1.2 Selection of Language Model |(LMs)

MuRIL and XLM-R both provide prominent results for the NER task for Multiconer-1 [17] and Hiner [24] dataset. Based on the performance of these two LMs on Hindi data, these models have been selected for the experiment.

1. **google/MuRIL-large-cased (MuRIL) [15]** The MuRIL model is trained on a large number of Indian text corpora, and it is augmented with both translated and transliterated document pairs to provide supervised cross-lingual signals in training. This allows the model to perform well on cross-lingual tasks and outperform multilingual BERT (mBERT) in the XTREME benchmark [31], which is a challenging cross-lingual benchmark for natural language processing tasks. Essentially, MuRIL is better at understanding and processing text in multiple languages, including Indian languages.

2. **XLM-Roberta-large (XLM-R) [16]** A masked language model based on the Transformer architecture, pre-trained on a diverse set of 100 languages and trained using over two terabytes of filtered CommonCrawl data, including Hindi.

The characteristics of each of them are organized in Table 3 for comparison.

| Base Model | XLM-Roberta | MuRIL |
|---|---|---|
| **Parameters** | 355M | 236M |
| **Hidden layers** | 24 | 24 |
| **Vocabulary** | 250002 | 197285 |
| **Language** | 100 | 17 |
| **Dataset used** | CommonCrawl, Wikipedia, Oscar | CommomCrawl, Wikipedia |

Table 3: XLM-R and MuRIL comparison on characteristics

### 4.1.3 General Architecture for NER

In Fig. 3, details of the architecture of NER have been shown. It consists of four steps: pre-processing, context encoding, classification layer and post-processing.

1. **Pre-processing** NER module uses a tokenizer to convert the raw text into numerical representations that LMs can process. This is typically done by applying a tokenizer, which breaks the text into tokens (words or subwords), and padding applies if the length of tokenized sentences is less than the tokenizer max length. Also, truncation can be applied if the length of tokenized sentences is greater than the tokenizer's max length. Thereafter, the tokenizer adds some special tokens. Finally, all tokenized sentences are mapped to unique integer IDs based on the vocab of the tokenizer. Tokenized text is given to a neural network for encoding in which an attention mask is used to indicate tokens that are part of the input sentence, and the rest of the padding can be ignored. The tokenizer of the MuRIL Large model is based on WordPiece [32] tokenizer while the tokenizer of XLM-R [16] is based on sentence-piece tokenizer[33].

   After tokenization for the alignment of labels with tokenised sentences, we expand labels. This expansion is done by assigning the same labels to each token, using the word assigned from which the token originated. An example of label alignment is shown in Fig. 4.1.1.

| Experimental Setup and Hyper-parameters | MuRIL | XLM-R |
|---|---|---|
| Baseline language model | google/MuRIL-large-cased | XLM-Roberta-large |
| Hidden size for language model | 1024 | 1024 |
| Classification layer | Linear layer with cross-entropy loss function | Linear layer with cross-entropy loss function |
| Learning rate for language models | 5e-06 | 5e-06 |
| Learning rate for the classification layer | 5e-06 | 5e-06 |
| training epochs | 20 | 20 |
| Batch size | 64 | 64 |
| Model_Max_Len for tokenizer | 512 | 512 |
| Dropout rate | 0.1 | 0.1 |
| Seed | 10 | 5 |
| Activation function | Relu | Relu |
| Optimizer | AdamW | AdamW |

Table 4: Experimental Setup and Hyperparameters for best model of MuRIL and XLM-R

2. **Context Encoder layer (MuRIL and XLM-R in this work)** The tokenizer output is fed into the selected LM, which extracts features based on input context information as a vector of size 1024 for each token. Transformer-based encoders encode the input sentences into the deep features vector for each token of the input, and these features have contextual information, which is learned by taking the attention of each token of the input.

3. **Classification Layer**

   MuRIL and XLM-R generate feature embeddings for each token of size 1024, so we map these embeddings to a vector of size 14, as 13 labels are provided in the original dataset and an additional one for augmented data. This vector is also called the logits vector. After generating logits, we first normalized with softmax and then applied cross-entropy for computing loss. The combined Formula for both functions is as follows.

   $$loss = -\sum_{c=1}^{C}[\log \frac{exp(x_c)}{[\sum_{i=1}^{C} exp(x_i)]} y_c]$$

   Where x is logits vector containing probabilities corresponding to each label of the input sentence, y is the vector of ground truth (1 or 0 for each token present in the input sentence), C is the number of classes (unique labels).

   During testing, we calculate labels by finding the index of the max value at logits.

4. **Post Processing** Once the model has generated the labels, they are adjusted to align with the original sentence. This process is the opposite of the label alignment discussed in the pre-processing section. In this case, the labels for all tokens that represent the first tokens of their original word are considered the generated labels.

   The realignment of labels ensures that the generated labels accurately correspond to the words in the original sentence.

### 4.1.4 Experimental setup

The best score was obtained using XLM-R as the model, but the performance of MuRIL was also comparable to that of XLM-R. The experiments were conducted using batch sizes of 16 and 64, with learning rates of 1e-06, 5e-06, and 1e-05, as well as various random seed values. The specific details regarding the experimental setup and hyperparameters for the best model in both configurations are described in Table 4.1.3. Implementation of the NER module with the Cross entropy classification layer is done by xxxtokenclassificaion class defined in [34], where xxx refers to the selected model. The best model was selected based on the best score achieved by the model upon validation data.

## 4.2 Finetunning Llama2-7B with and without RA

### 4.2.1 Selection of Large Language Models (LLMs)

We have chosen the Llama2-7B-chat-hf for fine-tuning because it is open source and lightweight among the Llama2 series. All characteristics of llama2-7B-chat-hf are similar to llama2-70B-chat-hf 5 except that it has 7B billion parameters [36].

### 4.2.2 Retrieval Augmentation

In this strategy, RA is carried out in a similar manner as described in section 4.1.1. We have used sentence-level searches for augmentation.

### 4.2.3 Data Pre-processing

Data preparation for fine-tuning involves five elements. The first element is the system prompt for the base model, followed by instructions for the named entity recognition (NER) task. The third element is the sentence for which NER is required, while the fourth element is a similar context of the sentence (retrieved augmented data). Finally, the last element is the predicted output in dictionary format. For example: "Entity1: Type of Entity1, Entity2: Type of Entity2, ...". The format of the input sentence is shown in Fig. 6.

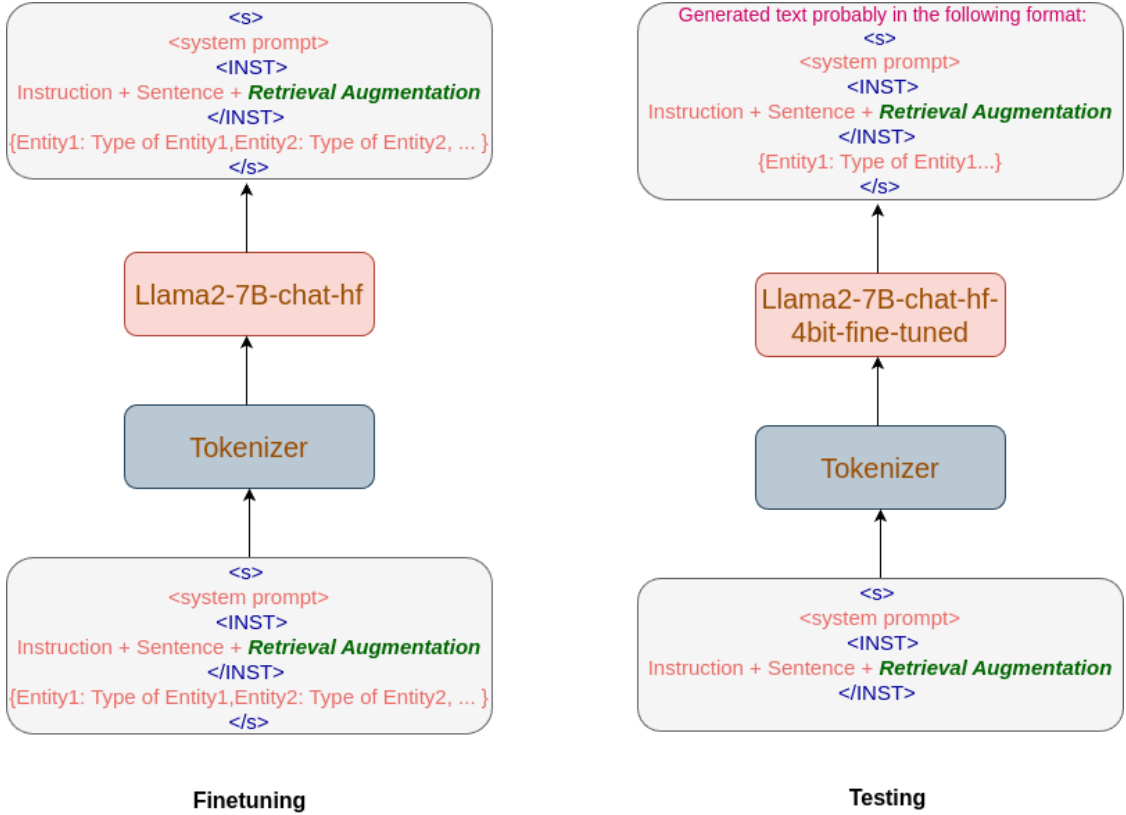

Figure 6: Architecture of LLM for NER prediction with and without RA

### 4.2.4 General Architecture

The overall architecture of RAFT is shown in Fig. 6. The maximum length of an input is set to 800, so retrieval augmentation is done accordingly. The input to the model includes a specific format consisting of a system prompt, instructions for the NER task, the sentence for which the model performs NER, retrieval augmented data, and output format in a dictionary form where keys represent entities and values represent their entity types. An example of the input format is: "<s><system prompt> <INST> Instruction + Sentence + Retrieval Augmentation </INST> Entity1: Type of Entity1, Entity2: Type of Entity2, ... </s>" System prompts and instructional guidelines help shape language model behaviour. Prompt tuning is a method for training and updating new prompt tokens in a pre-trained model. Causal language modelling has been used for fine-tuning. Causal language modelling involves predicting the next token in a sequence of tokens. During the testing phase, the input is provided to the model without an output format, and the model is expected to generate the output. We have also fine-tuned the model without retrieval augmentation. The only difference in this procedure is that retrieval augmentation of the input sentence is not done during input text formation.

### 4.2.5 Post Processing

In the post-processing step, we extracted the output in dictionary format from the generated text using regular expressions.

### 4.2.6 Experimental setup

We conducted experiments using batch sizes of 4 and 8, with learning rates of 2e-04, 1e-04, 3e-05, and 5e-05 and with max sequence length of 820. The models were fine-tuned for 2 epochs on our training data. Our best result was achieved with a learning rate of 5e-05 and a batch size of 8. We also observed that the model did not optimize for the task with learning rates of 2e-04 and 1e-04. For efficient training, QLoRA procedure has been utilised. QLoRA [39] is a finetuning method that involves quantizing a model to 4-bits and introducing a set of low-rank adaptation (LoRA) weights to the model. These LoRA weights are then tuned alongside the quantized weights to improve the model's performance." QLoRA is supported by the State-of-the-art Parameter-Efficient Fine-Tuning (PEFT) library by Hugging Face [34]. The additional parameters for QLoRA are: Attention dimension: 64, Scaling parameter: 16, Dropout probability: 0.1, 4-bit precision base model loading: True, Data type for 4-bit base models: "float16", Quantization type: "nf4". Implementation of the module with the Cross entropy classification layer is done by AutoModelForCausalLM class defined in [34].

### 4.3 NER Prediction with RAG-based Few-shot Prompting using Chain of Thoughts strategy with Generative Models

An overview of few-shot prompting using chain of thoughts strategy has been shown in Fig. 7 This section is further divided into the following subsections.

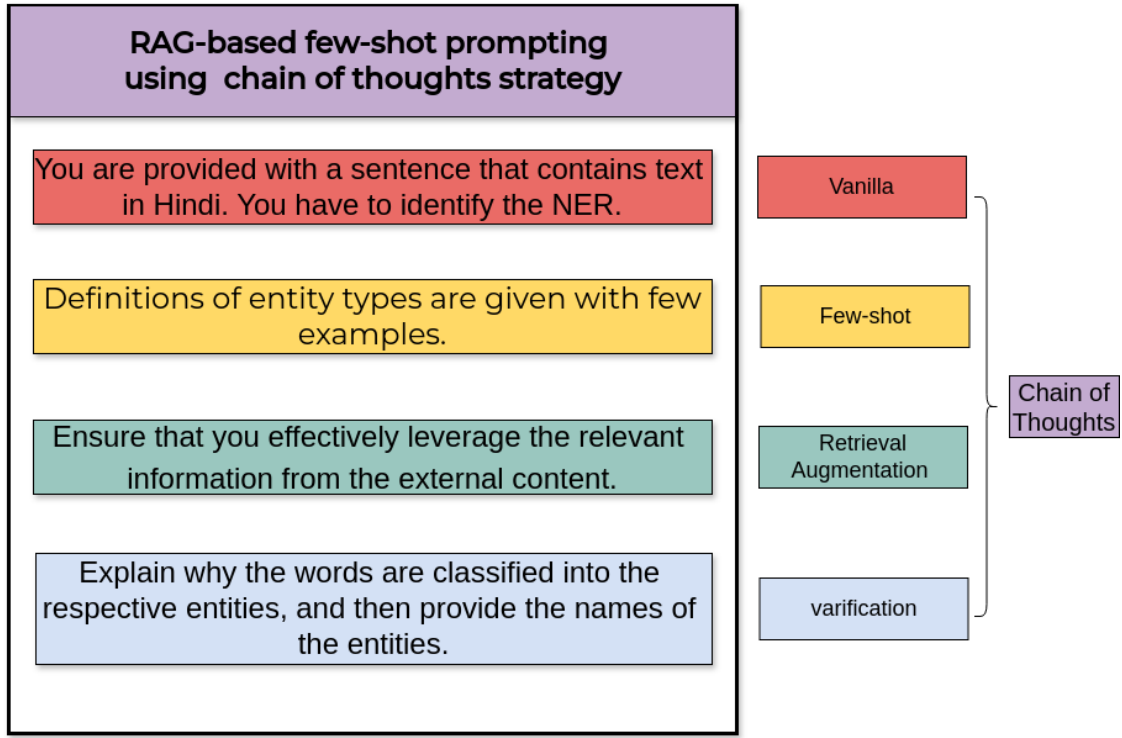

Figure 7: General architecture of RAG-based few-shot prompting using the chain of thoughts strategy.

#### 4.3.1 Selection of Large Language Models (LLMs)

Large Language Models (LLMs) have demonstrated significant potential as highly capable AI assistants. They excel in complex reasoning tasks requiring expert knowledge across various fields, including specialized domains like programming and creative writing. They facilitate interaction with humans through intuitive chat interfaces, leading to rapid and widespread adoption among the general public. We have selected four LLMs for our task: Llama2-7B, Llama2-70B, Llama3-70B) and GPT3.5 turbo.

1. **Llama-2-70b-chat-hf (Llama2-70B) and Llama-3-70B-Instruct (Llama3-70B) [36]** Llama2 is trained with 2 trillion tokens and has 70 billion parameters. It's a fine-tuned version optimized for dialogue use cases and improved with RLHF. It's open source and supports a context size of 4000 tokens. Llama2 performed the best among the Llama 2 series. Llama3 is trained on 15 trillion tokens with 70 billion parameters. Llama3 performed better compared to Llama2 in various tasks. A comparison of the characteristics of Llama2, Llama3, and GPT3.5-turbo is shown in Table 5.

2. **GPT3.5-turbo** It is an optimized version of GPT-3 [35], known as GPT-3.5 Turbo, which has been trained to understand natural language. These models are capable of comprehending and generating natural language as well as code. They have been optimized for chat using the Chat Completions API but are also suitable for non-chat tasks. The model supports multiple languages, including Hindi, and has a context window of 16,385 tokens. A context window refers to the textual range around a target token that a large language model (LLM) can process at the time the information is generated. It is not an open-source service, but it offers multiple paid versions for different usage options [27].

| Base Model | Llama 2-Chat (70B) | Llama 3-Chat (70B) | GPT3.5-turbo |
|---|---|---|---|
| **Open source** | Yes | Yes | No |
| **Context Window size** | 4096 tokens | 8192 tokens | 16,385 tokens |
| **Training Data (Tokens)** | 2.2 trillions | 15 trillions | 200 Billions |
| **Parameters** | 70B | 70B | - |
| **Language** | multi-lingual | multi-lingual | multi-lingual |

Table 5: Llama 2-Chat (70B), Llama 3-Chat (70B) and GPT3 comparison on characteristics

#### 4.3.2 Retrieval Augmentation

In this scenario, RA is carried out in a similar manner as described in section 4.1.1. However, there are some differences. Since there is no training involved in this case, iterative retrieval is not performed. Additionally, RA only conducts sentence-level searches. The process involves augmenting the original sentence by concatenating the retrieved cleaned paragraphs. We clean the paragraph by removing special symbols so that it can align with the prompt for LLMs.

#### 4.3.3 Few-shot Prompting

Based on the findings of different subtasks in [35, 29, 28], we employed few-shot prompting and chain-of-thought based on the results of some examples for this task. Our experiments with the methods outlined in [29] also validate the use of few-shot prompting. An overview of the prompts with RA is depicted in Fig. 7.

#### 4.3.4 Post Processing

GPT-3.5-turbo generates formatted results as required; however, Llama3 and Llama2 sometimes produce unstructured and random results. Therefore, we categorize patterns of the generated results and heuristically extract the NER tags from these patterns.

#### 4.3.5 Experimental setup

Llama2 and Llama3 were implemented with together AI API interface [38] where the maximum length of the prompt with RA context is set to 3100 for Llama2, 7680 for Llama3. Twenty thousand examples of the test data are taken for experiments with Llama2 and Llama3. gpt3.5 turbo model implemented by open AI framework [27]. The maximum length of the prompt with RA context is set to 16385 for gpt3.5 turbo, which is also the default context window size. Due to the cost effect, only 1000 test examples were taken with gpt3.5 for the experiment.

## 5 Evaluation metrics

For evaluation, strict F1[3] score applied. Strict evaluation matches both boundaries of an Entity (Beginning and Ending) and types ( Location, Person etc.) [37].

It can be computed using the following formula:

$$StrictF1Score = 2 \times \frac{\text{Strict Precision} \times \text{Strict Recall}}{\text{Strict Precision} + \text{Strict Recall}}$$

---
[3]https://github.com/MantisAI/nervaluate

We have taken the Macro average of all labels' (excluding "other") strict F1 scores for overall results.

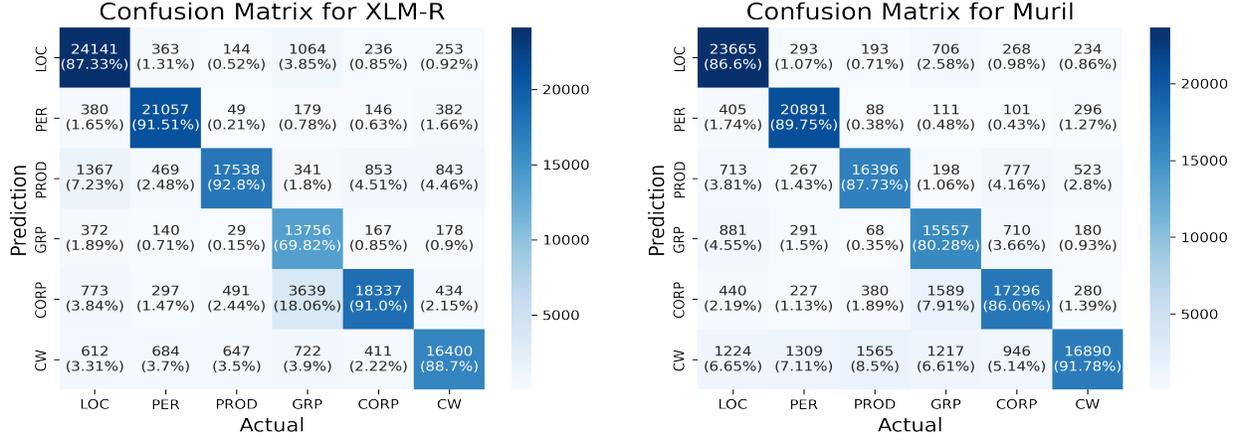

Figure 8: The results of the best models utilizing RA, XLM-R and MuRIL, are displayed in the Confusion Matrix.

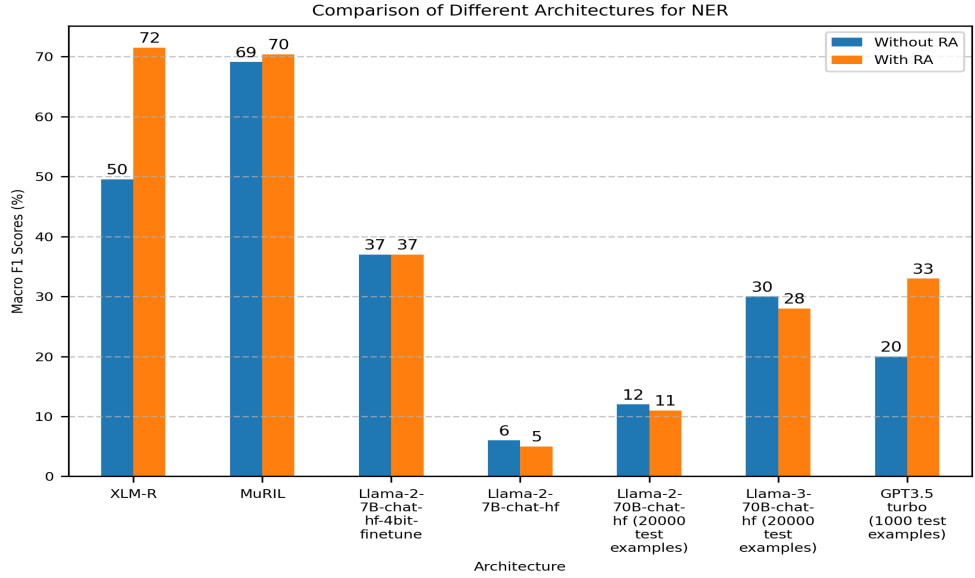

Figure 9: Figure shows the comparison between results with and without RA for language models.

# 6 Results and Discussion

## 6.1 Results with Llama2-7B-4bit-fine-tune

**Comparison between llama2-7B-chat-hf and fine-tuned model Llama2-7B-4bit-fine-tune (with and without RA)** After fine-tuning, the results improved in terms of accuracy and output format. The base model's macro F1-score is 6 without RA and 5 with RA. We also created some regular expressions to extract information from the generated output. However, after fine-tuning the model, its macro F1-score increased to 37 in both cases, with and without RA. The generated output has a fixed format (dictionary) for most sentences.

**Comparison between with RA and without RA using Llama2-7B-4bit-fine-tune** Fine-tuned models with llama2-7B-chat-hf using both the strategy (with and without RA) produce the same macro F1-score of 37.

| Architecture | Without RA (%) | With RA (%) |
|---|---|---|
| XLM-R | 49.55 | 71.50 |
| MuRIL | 69.08 | 70.40 |
| Llama-2-7B-chat-hf-4bit-finetune | 37.00 | 37.00 |
| Llama-2-7B-chat-hf | 6.00 | 5.00 |
| Llama-2-70B-chat-hf (20000 test examples) | 12.00 | 11.00 |
| Llama-3-70B-chat-hf (20000 test examples) | 30.00 | 28.00 |
| GPT3.5 turbo (1000 test examples) | 20.00 | 33.00 |
| GLiNER [40] | 27.80 | - |
| GPT3.5 turbo [41] | 27.30 | - |

Table 6: Comparison of Different Architectures for NER

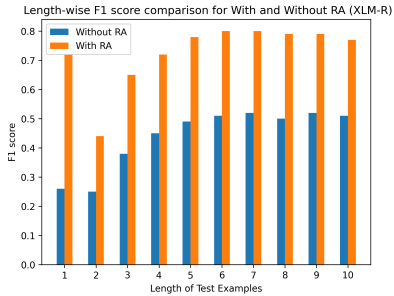

Figure 10: Comparing length-wise F1 scores with and without RA using XLM-R

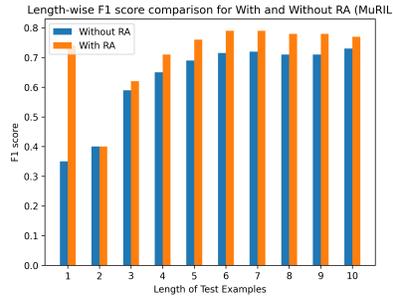

Figure 11: Comparing length-wise F1 scores with and without RA using MuRIL

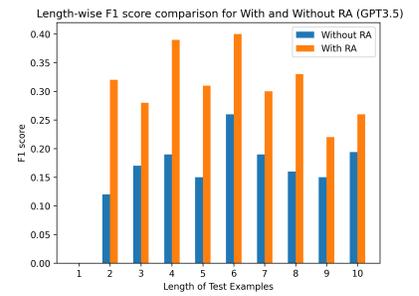

Figure 12: Comparing length-wise F1 scores with and without RA using GPT3.5

**Comparison of Llama2-7B-4bit-fine-tune with other Fine-tuned Models (Muril and XLM-R) with RA** Fine-tuned transformer-encoder-based models like Muril and XLM-R produce better results compared to fine-tuned transformer-decoder-based models like llama2-7B-chat-hf. The results in Table 6 and Fig. 9 depict that the Macro F1-score of XLM-R is 72, whereas the score for llama2-7B-chat-hf is 37.

**Comparison of Llama2-7B-4bit-fine-tune with Generative Models** From the results shown in Table 6, we can infer that the fine-tuned llama2-7B-chat-hf model outperforms the state-of-the-art generative models like llama2-7B-chat-hf, llama2-70B-chat-hf, llama3-70B-chat-hf and GPT3.5 turbo. Due to resource constraints, we have compared only 20000 examples with llama2-70B-chat-hf, llama3-70B-chat-hf and only 1000 examples with the GPT3.5 turbo. However, the work of [41] scored 27.3 F1-score in our dataset. One other state-of-the-art multilingual NER generative model GLiNER [40] scored 27.8 F1-score. Hence, our fine-tuned llama2-7B-chat-hf outperforms the GLiNER and GPT-3.5-turbo.

### 6.2 Results with XLM-R and MuRIL

The result shown in Table 7 expresses improvement with RA. In each iteration, RA is done using sentence retrieval and entity retrieval based on the entity predicted by the previous iteration. In the first iteration, results obtained by data are augmented by only sentence retrieval as, at this stage, test data do not know about the entity. The results of the second iteration are obtained by data, which are augmented by both sentence retrieval and entity retrieval as, at this stage, test data have some knowledge of entities predicted in the previous iteration. Similarly, we obtain results for the third iteration. It was observed that improvement was negligible in the third iteration, and improvement was saturated after the third iteration.

In Table 8, the results without RA and results where RA is implemented with the entities which are generated without RA are shown. For MuRIL, the fourth iteration's result with RA is less than the result without RA. Hence, we used entities predicted by the model without RA (which has a 0.69 macro F1 score) for RA; it improves the macro F1 score by more than one percent. Results with XLM-R showed a good improvement over iterations, and around twenty-one percent improvement was achieved. This shows the effect of RA, which can help extract more relevant feature vectors.

| Pretrained Model | XLM-R | MuRIL |
|---|---|---|
| F1 Score (First iteration Retrieval Augmentation) | 0.6464 | 0.6401 |
| F1 Score (Second iteration Retrieval Augmentation) | 0.7151 | 0.6889 |
| F1 Score (Third iteration Retrieval Augmentation) | 0.7153 | 0.6895 |
| F1 Score (Fourth iteration Retrieval Augmentation) | 0.7153 | 0.6895 |

Table 7: The Macro F1 score of each iteration with RA is shown. At the fourth iteration, the F1 score is saturated.

| Pretrained Model | XLM-R | MuRIL |
|---|---|---|
| F1 Score (Without Retrieval Augmentation) | 0.4955 | 0.6908 |
| F1 Score (Retrieval Augmentation with (previous best model predicted entities ) | 0.6701 | 0.7041 |

Table 8: Macro F1 score without RA for test data are shown in row number (row) 1. The last row shows the F1 score where prediction is generated with the RA with the entities which are generated without RA.

While MuRIL exhibits equivalent performance in both cases, XLM-R captures knowledge more effectively when assisted by RA. This observation shows that MuRIL has a better knowledge of the Hindi Multiconer1 dataset, and RA improves XLM-R's capabilities.

Figures 10 and 12 depict a comprehensive comparison of the length-wise F1-score with and without RA for the XLM-R and MuRIL. For simplicity, in both the figures, we have shown only those examples whose lengths are less than 15. It has been noticed that RA has given improvements for small-length sentences and shows that the difference between the scores decreases with increasing the length of sentences. It indicates that RA performs well for low-context examples. Table 7 and Fig. 9 verify that the improvement done by XLM-R is notable. An example from the test set shown in Fig. 1 depicts the effect of RA by the MuRIL model. Details of the results in numerical form for XLM-R and MuRIL are tabulated in Table 9 and 10.

The confusion matrix of the best prediction of both models is shown in Fig. 8. It infers that MuRIL is more biased in predicting class Creative-Work than XLM-R, as more than five percent of other labels predicted Creative-Work by the MuRIL. Observing that the class "Group" is predicted as "Corporation" approximately 18 percent of the time by XLM-R, compared to around seven percent for MuRIL, we can deduce that the XLM-R setup exhibits more confusion when predicting the "Group" class as "Corporation.". At the same time, MuRIL handles this ambiguity to some extent.

### 6.3 Results comparison with Llama2-70B, Llama3-70B, and GPT3.5 turbo using Few-shot prompting

Results with Llama2, Llama3 and GPT3.5-turbo are shown in Table 6. Llama3-70B and GPT3.5-turbo perform better than llama2-70B in both the cases with RA and without RA. Llama3-70B performs better than GPT3.5-turbo if it is without RA. However, GPT3.5-turbo adopts RA as with RA GPT3.5-turbo outperforms Llama3-70B. Moreover, GPT3.5-turbo predicts in the required format, and Llama2 predict in multiple formats, which we convert in the required format with heuristic approaches after analyzing the pattern of generated outputs.

**Comparison between RA and without RA** Results with RA and without RA tabulated in Table 9. Results with RA perform better for gpt3.5-turbo. However, for Llama2-70B and Llama3-70B, results with the were inferior to those without RA. There can be multiple reasons for this. First can be the context window size of Llama2 and Llama3, which is less than the context window size of gpt3.5-turbo; refer to Table 5. A longer context window length allows for more data augmentation. The second reason is that Llama2-70B generates data in multiple formats. Hence, Extracting named entities from generated data can be complex.

**Comparison between fine-tuned transformer-encoder based models (XLM-R and MuRIL) and generative models (Llama2-70B, Llama3-70B and GPT3.5-turbo) based few-shot prompting** Transformer-encoder-based models utilized all training examples for the training and therefore performed better compared to LLMs. However, LLMs utilized only a few examples during prompting in a few-shot manner, which shows a better possibility in the case of less data availability. The results of Transformer-encoder-

based models were better because these models are fine-tuned. The other reason for the low performance of LLMs is the test data format. Many predictions are correct based on manual perception, but the test data has a different format.

### 6.4 Comparison of Resources requirement

Due to the fact that the parameters of Generative models are around 100 to 1000 times those of Transformer encoder-based models, Transformer encoder-based models (XLM-R with 355M parameters and MuRIL with 236 M parameters) require less RAM and time for prediction than Generative models (Llama2-7B , Llama3-70B and gpt3.5 turbo). Based on the overall analysis, we can infer that for the NER task, Transformer encoder-based models (such as XLM-R and MuRIL) will be more efficient in terms of resource requirements and time complexity if we have the proper dataset. If proper datasets are not available, few-shot prompting can still work to some extent.

## 7 Conclusion

We have implemented the RA approach and fine-tuned different models, MuRIL, XLM-R and Llama2-7B, to perform the task. These models were applied to the augmented Hindi NER dataset, and the data for augmentation was retrieved from Wikipedia. This work showed that if an LM does not extract relevant feature vectors of the example (input sentence for the NER task), retrieval augmentation can help the LM extract relevant feature vectors.

Label-wise and context-wise analysis was made, which helped to understand the improvements using the data augmentation approach. Our comprehensive study demonstrates the efficacy of RA with Wikipedia in enhancing NER performance. Notably, the improvement achieved with Llama2-7B-chat-hf and XLM-R is significantly more prominent. While MuRIL shows a relatively smaller improvement. MuRIL performance without RA is already comparable to XLM-R with RA; therefore, it can be inferred that if an LM already possesses a strong capability to comprehend low-context examples, RA may not be significantly effective.

Apart from the fine-tuning experiment with the generative models has been done where we have generated named entities for a sentence with and without RA. Where GPT3.5-turbo showed improvement with RA; however, Llama-2-70B and Llama-3-70B do not.

The results of this work suggest that RA can be helpful in low-context and low-resource scenarios.

**Declaration of competing interest**

This research did not receive any specific grant from funding agencies in the public, commercial, or not-for-profit sectors.

| Model | Length-wise Examples | F1 score (With KB) | | | | | | | F1 score (Without KB) | | | | | | |
|---|---|---|---|---|---|---|---|---|---|---|---|---|---|---|---|
| | | CW | PROD | PER | LOC | CORP | GRP | Macro F1 | CW | PROD | PER | LOC | CORP | GRP | Macro F1 |
| XLM-R | 1 | 0.69 | 0.83 | 0.8 | 0.76 | 0.88 | 0.38 | 0.72 | 0.25 | 0.28 | 0 | 0.82 | 0.05 | 0.15 | 0.26 |
| | 2 | 0.43 | 0.64 | 0.3 | 0.5 | 0.59 | 0.19 | 0.44 | 0.26 | 0.32 | 0.14 | 0.43 | 0.22 | 0.14 | 0.25 |
| | 3 | 0.56 | 0.76 | 0.78 | 0.65 | 0.72 | 0.44 | 0.65 | 0.29 | 0.34 | 0.43 | 0.47 | 0.44 | 0.32 | 0.38 |
| | 4 | 0.69 | 0.79 | 0.81 | 0.71 | 0.77 | 0.57 | 0.72 | 0.38 | 0.40 | 0.46 | 0.52 | 0.49 | 0.43 | 0.45 |
| | 5 | 0.75 | 0.84 | 0.86 | 0.75 | 0.83 | 0.62 | 0.78 | 0.43 | 0.42 | 0.55 | 0.56 | 0.54 | 0.47 | 0.49 |
| | 6 | 0.8 | 0.83 | 0.86 | 0.79 | 0.84 | 0.65 | 0.8 | 0.45 | 0.47 | 0.53 | 0.58 | 0.55 | 0.49 | 0.51 |
| | 7 | 0.8 | 0.85 | 0.88 | 0.8 | 0.83 | 0.66 | 0.8 | 0.49 | 0.45 | 0.55 | 0.57 | 0.54 | 0.50 | 0.52 |
| | 8 | 0.8 | 0.8 | 0.84 | 0.79 | 0.84 | 0.67 | 0.79 | 0.49 | 0.47 | 0.49 | 0.54 | 0.55 | 0.46 | 0.50 |
| | 9 | 0.79 | 0.74 | 0.84 | 0.79 | 0.84 | 0.73 | 0.79 | 0.48 | 0.45 | 0.58 | 0.52 | 0.57 | 0.52 | 0.52 |
| | 10 | 0.72 | 0.75 | 0.86 | 0.77 | 0.82 | 0.7 | 0.77 | 0.45 | 0.46 | 0.59 | 0.49 | 0.54 | 0.52 | 0.51 |
| | 11 | 0.69 | 0.67 | 0.87 | 0.77 | 0.79 | 0.68 | 0.75 | 0.43 | 0.51 | 0.63 | 0.47 | 0.52 | 0.51 | 0.51 |
| | 12 | 0.63 | 0.65 | 0.86 | 0.77 | 0.74 | 0.73 | 0.73 | 0.45 | 0.45 | 0.58 | 0.51 | 0.48 | 0.64 | 0.52 |
| | 13 | 0.64 | 0.67 | 0.83 | 0.76 | 0.77 | 0.77 | 0.74 | 0.50 | 0.47 | 0.57 | 0.52 | 0.54 | 0.71 | 0.55 |
| | 14 | 0.6 | 0.59 | 0.84 | 0.73 | 0.83 | 0.77 | 0.73 | 0.44 | 0.48 | 0.55 | 0.52 | 0.57 | 0.67 | 0.53 |
| | 15 | 0.57 | 0.63 | 0.81 | 0.8 | 0.8 | 0.64 | 0.71 | 0.44 | 0.45 | 0.62 | 0.52 | 0.54 | 0.58 | 0.52 |
| | All Examples | **0.68** | **0.73** | **0.81** | **0.71** | **0.76** | **0.59** | **0.72** | **0.39** | **0.43** | **0.55** | **0.50** | **0.54** | **0.53** | **0.49** |

Table 9: Detail of all results of XLM-R. The F1 score of each label for both cases (with and without the knowledge base) is tabulated. Along with macro F1 score for all labels for test data, length-wise F1 score for examples whose lengths are less than 15 are given.

| Model | Length-wise Examples | F1 score (With KB) | | | | | | | F1 score (Without KB) | | | | | | |
|---|---|---|---|---|---|---|---|---|---|---|---|---|---|---|---|
| | | CW | PROD | PER | LOC | CORP | GRP | Macro f1 | CW | PROD | PER | LOC | CORP | GRP | Macro f1 |
| MuRIL | 1 | 0.67 | 0.76 | 1 | 0.86 | 0.65 | 0.49 | 0.74 | 0.41 | 0.41 | 0 | 0.83 | 0.45 | 0.05 | 0.35 |
| | 2 | 0.45 | 0.58 | 0.35 | 0.33 | 0.42 | 0.27 | 0.4 | 0.48 | 0.54 | 0.22 | 0.65 | 0.36 | 0.15 | 0.4 |
| | 3 | 0.54 | 0.74 | 0.76 | 0.53 | 0.62 | 0.51 | 0.62 | 0.53 | 0.68 | 0.7 | 0.59 | 0.62 | 0.46 | 0.59 |
| | 4 | 0.72 | 0.75 | 0.8 | 0.64 | 0.73 | 0.64 | 0.71 | 0.62 | 0.68 | 0.68 | 0.67 | 0.67 | 0.59 | 0.65 |
| | 5 | 0.77 | 0.79 | 0.82 | 0.72 | 0.78 | 0.71 | 0.76 | 0.64 | 0.69 | 0.73 | 0.72 | 0.72 | 0.64 | 0.69 |
| | 6 | 0.79 | 0.8 | 0.85 | 0.75 | 0.78 | 0.75 | 0.79 | 0.67 | 0.73 | 0.75 | 0.73 | 0.73 | 0.68 | 0.715 |
| | 7 | 0.8 | 0.79 | 0.86 | 0.75 | 0.77 | 0.76 | 0.79 | 0.64 | 0.74 | 0.79 | 0.75 | 0.74 | 0.66 | 0.72 |
| | 8 | 0.82 | 0.76 | 0.82 | 0.73 | 0.77 | 0.77 | 0.78 | 0.64 | 0.71 | 0.77 | 0.75 | 0.73 | 0.66 | 0.71 |
| | 9 | 0.79 | 0.72 | 0.86 | 0.72 | 0.78 | 0.8 | 0.78 | 0.65 | 0.66 | 0.82 | 0.76 | 0.72 | 0.7 | 0.71 |
| | 10 | 0.74 | 0.75 | 0.86 | 0.73 | 0.79 | 0.76 | 0.77 | 0.67 | 0.72 | 0.84 | 0.72 | 0.77 | 0.71 | 0.73 |
| | 11 | 0.73 | 0.7 | 0.86 | 0.77 | 0.71 | 0.79 | 0.76 | 0.64 | 0.72 | 0.88 | 0.77 | 0.76 | 0.72 | 0.74 |
| | 12 | 0.69 | 0.74 | 0.83 | 0.75 | 0.73 | 0.79 | 0.75 | 0.67 | 0.72 | 0.88 | 0.77 | 0.79 | 0.77 | 0.76 |
| | 13 | 0.67 | 0.71 | 0.87 | 0.77 | 0.73 | 0.84 | 0.77 | 0.71 | 0.74 | 0.9 | 0.78 | 0.8 | 0.84 | 0.79 |
| | 14 | 0.63 | 0.66 | 0.84 | 0.8 | 0.79 | 0.81 | 0.76 | 0.67 | 0.67 | 0.86 | 0.79 | 0.82 | 0.79 | 0.76 |
| | 15 | 0.68 | 0.71 | 0.86 | 0.75 | 0.77 | 0.74 | 0.75 | 0.69 | 0.68 | 0.89 | 0.78 | 0.77 | 0.79 | 0.76 |
| | All Examples | 0.69 | 0.71 | 0.8 | 0.65 | 0.7 | 0.67 | 0.7 | 0.56 | 0.69 | 0.77 | 0.67 | 0.71 | 0.71 | 0.69 |

Table 10: Detail of all results of MuRIL. The F1 score of each label for both cases (with and without the knowledge base) is tabulated. Along with macro F1 score for all labels for test data also, length-wise F1 score for examples whose lengths are less than 15 are given.